\documentclass{bmvc2k}
\usepackage{float}
\usepackage{enumitem}

\title{Optimal Transport Based Generative Autoencoders}

\addauthor{Oliver Zhang}{oliverzhang42@gmail.com}{1}
\addauthor{Ruei-Sung Lin}{rueisung@gmail.com}{2}
\addauthor{Yuchuan Gou}{plain1994@gmail.com}{2}

\addinstitution{
 Proof School\\
 973 Mission Street\\
 San Francisco, CA, USA
}
\addinstitution{
 Ping An Technology\\
 3000 El Camino Real\\
 5 Palo Alto Square, Suite 150\\
 Palo Alto, CA, USA
}

\runninghead{Zhang, Lin, Gou}{Optimal Transport Based Generative Autoencoders}

\def\etal{\emph{et al}\bmvaOneDot}

\begin{document}

\maketitle

\begin{abstract}
The field of deep generative modeling is dominated by generative adversarial networks (GANs). However, the training of GANs often lacks stability, fails to converge, and suffers from model collapse. It takes an assortment of tricks to solve these problems, which may be difficult to understand for those seeking to apply generative modeling. Instead, we propose two novel generative autoencoders, AE-OTtrans and AE-OTgen, which rely on optimal transport instead of adversarial training. AE-OTtrans and AE-OTgen, unlike VAE and WAE, preserve the manifold of the data; they do not force the latent distribution to match a normal distribution, resulting in greater quality images. AE-OTtrans and AE-OTgen also produce images of higher diversity compared to their predecessor, AE-OT. We show that AE-OTtrans and AE-OTgen surpass GANs in the MNIST and FashionMNIST datasets. Furthermore, We show that AE-OTtrans and AE-OTgen do state of the art on the MNIST, FashionMNIST, and CelebA image sets comapred to other non-adversarial generative models.
\end{abstract}

\section{Introduction}
Deep generative modeling is the field of modeling high dimensional data distributions through different neural network architectures. It has widespread applications including text generation, data augmentation, and speech synthesis. The two approaches dominating the field are generative adversarial networks (GANs) developed by Goodfellow \etal [5] and generative autoencoders, the most prominent of which is variational autoencoders (VAEs) by Kingma and Welling [1]. GANs involve a minmax game between a generator and a discriminator, and training the GAN is often difficult, prone to exploding or vanishing gradients as well as mode collapse [6, 7, 8]. Variational Autoencoders, instead only require optimizing a simple minimization problem and thus are easier to train. Latent space density function mapping is at the heart of VAEs, and improvements to this density function mapping have been explored in two different ways, embodied by WAE by Tolstikhin \etal [2] and AEOT by Liu \etal [3]. WAE adds a regularizer term to force the latent space of the WAE to match a normal distribution. Theoretically, this makes generation of images much easier, as any vector sampled from a normal distribution should be familiar to the decoder. However, artificially regularizing the latent space to match a normal distribution lowers the quality of the image generation. AE-OT manages to maintain the shape of the latent space of the autoencoder, yet does not do a good job in mapping noise vectors to the proper vectors. 

%We propose novel generative models, AE-OTtrans and AE-OTgen. These models are easier than GAN to train because it only involves a minimization problem. Like AE-OT and unlike VAE and WAE, they preserve the latent structure of the autoencoder and do not force the latent distribution to be a gaussian or other prior, helping image quality. AE-OTtrans and AE-OTgen apply optimal transport more directly than AE-OT, resulting in images of higher quality. Finally, AE-OTtrans and AE-OTgen apply optimal transport in slightly different ways; AE-OTtrans is more like a transporter and AE-OTgen is more like a generator.

We propose novel generative mapping algorithms OTtrans and OTgen. These algorithms leverage optimal transport to train deep neural networks to generate samples from lower dimensional data distributions. To generate high dimensional data, we first apply an autoencoder to the high dimensional data and then apply OTtrans and OTgen to sample from the latent distribution of the autoencoder. Depending on which algorithm we use, we call this two step procedure either AE-OTtrans or AE-OTgen. 

AE-OTtrans and AE-OTgen have superior performance compared to GAN in lower complexity datasets including MNIST and FashionMNIST. This is largely because optimal transport mitigates the model collapse problem in simpler datasets. Furthermore, AE-OTtrans and AE-OTgen are simpler to train and are more theoretically understood than GANs. Compared to non-adversarial generative models, AE-OTtrans and AE-OTgen generate high quality images and interpolations. As opposed to VAE and WAE, AE-OTtrans and AE-OTgen preserve the latent structure of the autoencoder and do not force the latent distribution to be a gaussian or other prior. This provides more flexibility for the model, which translates into the generation of higher fidelity data. As opposed to latent space generator of AE-OT, OTtrans and OTgen do not need to train a discriminator and thus are able to deal with sparser datasets. As a result, AE-OTtrans and AE-OTgen are able to serve as much better generators than AE-OT.

\section{Related Works}
The field of deep generative modeling solves the problem of sampling from high dimensional probability distributions often lying on a much lower dimensional manifold. For instance, deep models for face generation are able to sample images from latent spaces of hundreds of thousands of pixels, far too large for traditional sampling techniques. The two methods most prominent in the field are GANs and VAEs, which both leverage the lower dimensional manifold. VAEs do this through an autoencoder, whereas GANs do this through adversarial training. Specifically, GANs pit a generator against a discriminator in a two player minmax zero sum game, in which the generator tries to generate images to fool a discriminator, and the discriminator tries to distinguish between the generated and real images. The discriminator eventually manages to learn which images are on the lower dimensional manifold, whereas the generator learns to generate images which are on the manifold. In practice GANs are able to model complex datasets, such as the CelebA dataset, producing much more samples of higher quality than non-adversarial generative modeling [10]. Yet achieving this performance is difficult, a variety of different tricks [6,7,8,9] as well as much trial and error with to carefully procure the correct hyper parameters. We restate that non-adversarial generative modeling is still valuable for its well understood behavior, ease of training, and superior performance on less complex data.

\subsection{Base VAE}
VAEs is an approach which only requires a single minimization optimization problem. It consists of an encoder $Q(Z|X): \mathcal{X} \to \mathcal{Z}$ and a decoder $P(X|Z): \mathcal{Z} \to \mathcal{X}$. The encoder takes an image and condenses it into a series of means and standard deviations which parameterize a multi-dimensional normal distribution in the latent space. A vector is sampled from this distribution and then put through the encoder. The probabilistic nature of the encoder forces the decoder to generalize to most points within the latent space. Then, to generate points, simply sample vectors from an $n$-dimensional normal distribution and feed it through the decoder. However, as noted before [2, 3], the images generated by VAE tend to be more blurry than real images. This is because VAE's stochastic training algorithm introduces some uncertainty to the autoencoder, which responds by blurring the image to minimize the mean squared loss. 

\subsection{WAE}
WAE is an improvement on the base VAE. It consists of a deterministic autoencoder with an added cost term forcing the latent distribution of the AE to be similar to a normal distribution. It is different from VAE in that VAE encodes a single point to a normal distribution whereas in WAE, the cumulative distribution of the whole batch is penalized to match a normal distribution. This enables WAE to have a much higher reconstruction quality. Furthermore, WAE can calculate the divergence between the latent distribution and the normal distribution in two ways. WAE-GAN does this with a GAN in the latent space, whereas WAE-MMD uses Maximum Mean Discrepancy (MMD). Because WAE-GAN uses adversarial training in the latent space, we compare our model to WAE-MMD instead. WAE-MMD inevitably performs worse than a vanilla autoencoder at reconstruction of images, as it needs to satisfy the MMD penalty. Furthermore, the difference between  image quality in reconstructed images and generated images further suggests that the latent space of WAE-MMD doesn't truly match a normal distribution. Thus, MMD is a suboptimal metric for regularizing the latent distribution.

\subsection{AE-OT}
AE-OT [3] pretrains an autoencoder, consisting of an encoder $Q(Z|X): \mathcal{X} \to \mathcal{Z}$ and a decoder $P(X|Z)$ on the data. Then, it trains a neural network $D$ to distinguish between real points in the latent distribution and noise generated from a prior $\mathcal{N}$. It does this by approximating the Kantorovich potential, where a higher Kantorovich potential corresponds to a point more likely to be from the real distribution and not the prior. To generate images, simply sample noise from the normal distribution. The network $D$ will map $z$ to $z + \nabla D(z)$, which should have a high Kantorovich potential and thus should be likely to be in the real latent distribution. The final image is then $P(z + \nabla D(z))$. This algorithm preserves the latent space of the autoencoder, and thus any reconstructed images are very sharp. 

However, AE-OT also has some flaws. In practice, training an optimal discriminator is extremely difficult due to the sparse nature of the dataset. Even when we reduce dimensionality to 64, the discriminator is unable to serve as a good generator.

\section{Our Generative Autoencoders}
In this paper, we follow [3,4] and propose a two step generative model framework for high dimensional data, such as images. First reduce the dimensionality of the high dimensional data by training an encoder $Q_\phi(Z|X): \mathcal{X} \to \mathcal{Z}$ and a decoder $P_\theta(X|Z): \mathcal{Z} \to \mathcal{X}$ in an autoencoder framework. Then, train a model $f_w:$ which maps from a noise distribution $\mathcal{N}$ to the latent distribution $\mathcal{Z}$. To generate data similar to $X$, simply sample $N$ from $\mathcal{N}$. The vectors $f_w(N)$ will approximate $Z$ and the decoded data $P_\theta(f_w(N))$ will approximate $X$. 

We propose two possible ways to do latent distribution mapping, named OTtrans and OTgen. As we will see, OTtrans is more similar to transporting from a distribution to another distribution whereas OTgen is more similar to generating points from a distribution. The corresponding image generative models with autoencoders are then named AE-OTtrans and AE-OTgen respectively.

\subsection{Transporter: OTtrans}
In OTtrans, we train a neural network to approximate optimal transport. Let $\mathcal{Z}$ be the distribution we want to generate from, and let $\mathcal{N}$ be our prior distribution, which will often be a noise distribution. First, sample vectors $Z \sim \mathcal{Z}$ and $N \sim \mathcal{N}$ with $|N| = |Z|$. Then calculate the optimal transport mapping denoted by $\sigma$, a bijection from $N$ to $Z$. This is time we use optimal transport in the OTtrans algorithm. Finally, train a neural network $f_w$ which approximates this mapping. Specifically, $f_w$ should attempt to map $n_i \in N$ to $\sigma(n_i) \in Z$.

\begin{figure}
    \centering
    \includegraphics[scale=0.2]{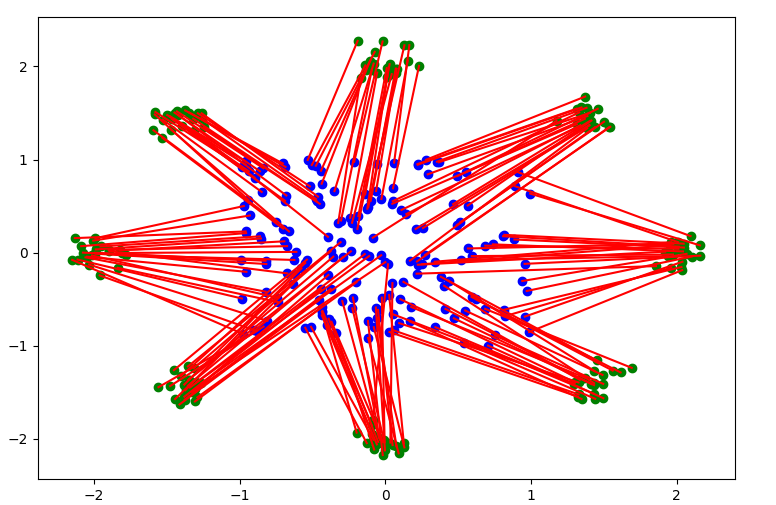}
    \caption{The OTtrans model. Noise vectors in blue, sampled from $\text{unif}(-1, 1)$, are optimally transported to the latent distribution in green. A neural network is then trained to learn this mapping.}
\end{figure}
\vspace{2mm}

\rule{0.9\linewidth}{0.1mm}

\vspace{0.5mm}
Algorithm 1: OTtrans

\vspace{-2mm}
\rule{0.9\linewidth}{0.1mm}

\vspace{0.5mm}
\textbf{Prerequistites:} Start by initializing encoder $Q_\phi(Z|X)$, decoder $P_\theta(X|Z)$ and transport 

neural network $f_w$. Let $c(a, b)$ be the corresponding squared cost between $a$ and $b$, and 

let $\mathcal{N}$ be the prior distribution.

\vspace{2mm}

1. Sample $X = \{x_1, x_2, ..., x_m\}$ from the training set and $N = \{n_1, n_2, ..., n_m\}$ from $\mathcal{N}$.

2. Encode $X$ to $Z=\{z_1, z_2, ..., z_m\}$ with $z_i = Q_\phi(x_i)$

3. Calculate the optimal transport map $\sigma$, a bijection from $N$ to $Z$.

\textbf{while} $w$ is not converged \textbf{do}:

{\addtolength{\leftskip}{5mm}

4. Randomly sample batch $\hat{N} \subset N$.

5. Calculate loss: 
\begin{equation}
\mathcal{L}(w) = \frac{1}{|\hat{N}|}\sum_{n_i \in \hat{N}}^{k}c(f_w(n_i), \sigma(n_i)))
\end{equation}

\hspace{4mm} 6. Update $w$ by using Adam to minimize loss.
}

\vspace{4mm}

AE-OTtrans is significantly different from AE-OT. Whereas AE-OT relies on a network to learn the Kantorovich Duals and approximates the optimal transport mapping, AE-OTtrans trains a network to learn the optimal transport mapping directly. Learning the mapping directly takes away any need to approximate or concern oneself with the model's first derivatives. The resulting model is more robust to sparse datasets and easier to train.

\subsection{Latent Space Generator: OTgen}
In OTgen, we train a neural network to generate points by using optimal transport to give "feedback" to the network. In contrast to the previous algorithm, we calculate optimal transport multiple times, at every iteration of the training step. Let $\mathcal{Z}$ be the latent distribution we want to generate from, and let $\mathcal{N}$ be the prior distribution. First, sample batch $Z \sim \mathcal{Z}$ and $N \sim \mathcal{N}$. Enumerate $Z = \{z_1, z_2, ..., z_k\}$ and $N = \{n_1, n_2, ..., n_k\}$. Then calculate the predictions $P = \{p_1, p_2, ..., p_k\}$ made by the network $f_w$, such that $p_i = f_w(n_i)$. 

We then use optimal transport to calculate "feedback" for each $p_i$. Find the optimal transport mapping on these predictions to get bijection $\sigma$ which maps $p_i$ to some $z_j \in Z$. Intuitively, $p_i$ should have been $\sigma(p_i)$. Finally, update $f_w$ based on this new optimal transport mapping. Specifically, $f_w$ should attempt to map $n_i$ to $\sigma(p_i) = \sigma(f_w(n_i))$.

\begin{figure}
    \centering
    \begin{minipage}[t]{0.45\textwidth}
        \centering
        \includegraphics[width=\textwidth]{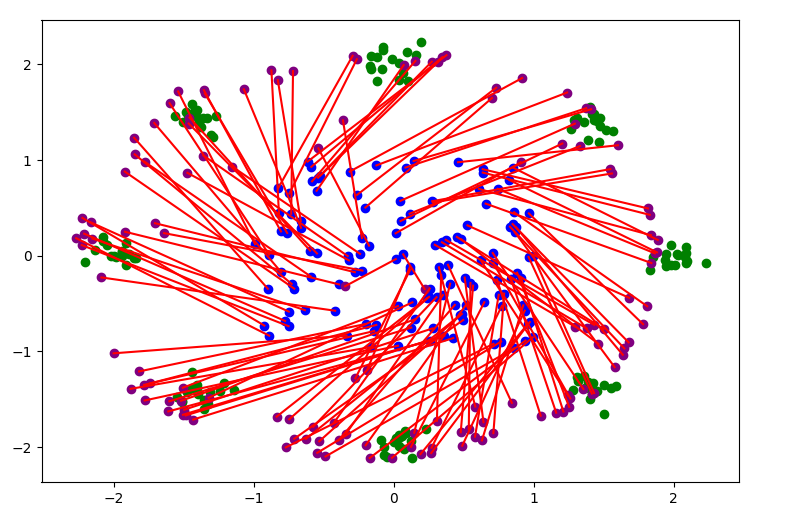} 
        \caption{Noise vectors in blue are sent to predictions in purple by $f_w$. Before the network can finish this training step, it requires feedback (figure 4).}
    \end{minipage}
    \hspace{4mm}
    \begin{minipage}[t]{0.45\textwidth}
        \centering
        \includegraphics[width=\textwidth]{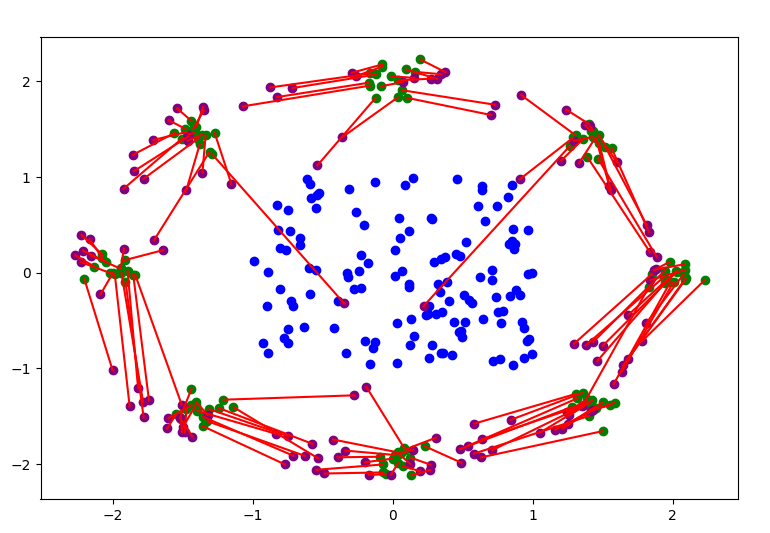}
        \caption{An optimal transport map $\sigma$ takes predictions in purple to the latent distribution in green. This shows $f_w$ where the purple points should have been and $f_w$ is updated accordingly.}
    \end{minipage}
\end{figure}

\noindent Finally, we also add an term $D(f_w(N), Z)$ weighted by a hyper parameter $\lambda$ to increase diversity. $D(f_w(N))$ takes the average distance between two generated latent vectors and compares it to the average distance between two genuine latent vectors. This forces the generated vectors to be, on average, as far apart as the genuine latent vectors. This is especially useful in a high dimensional latent space, and for many smaller dimensional latent spaces, $\lambda=0$ works well. Formally, it is calculated:

$$D(f_w, N, Z) = \left|\frac{1}{|N|}\left(\sum\limits_{n_1, n_2 \in N} f_w(n_1) - f_w(n_2) \right) - \frac{1}{|Z|}\left(\sum\limits_{z_1, z_2 \in Z} z_1 - z_2 \right) \right|_1$$

\vspace{5mm}

\rule{\linewidth}{0.1mm}

\vspace{0.5mm}
Algorithm 2: OTgen

\vspace{-2mm}
\rule{\linewidth}{0.1mm}

\vspace{0.5mm}
\textbf{Prerequistites:} Start by initializing encoder $Q_\phi(Z|X)$, decoder $P_\theta(X|Z)$, and generative 

neural network $f_w$. Let $c(a, b)$ be the corresponding squared cost between $a$ and $b$, let 

$\mathcal{N}$ be the prior distribution, and let $\lambda$ be the weight of the divergence term.

\vspace{2mm}

\textbf{while} $w$ is not converged \textbf{do}:

{\addtolength{\leftskip}{5mm}
1. Sample $X = \{x_1, x_2, ..., x_k\}$ from the training set and $N = \{n_1, n_2, ..., n_k\}$ from $\mathcal{N}$.

2. Encode $X$ to $Z = \{z_1, z_2, ..., z_k\}$ with $z_i = Q_\phi(x_i)$

3. Calculate the predictions $P = \{f_w(n_1)$, $f_w(n_2)$, ..., $f_w(n_k)\}$ made by $f_w$.

4. Calculate the optimal transport map $\sigma$, a bijection from the predictions $P$ to $Z$.

5. Calculate loss: 
\begin{equation}
\mathcal{L}(w) = \left( \frac{1}{k}\sum_{i=1}^{k}c(f_w(n_i), \sigma(f_w(n_i)))) \right) + \lambda \cdot D(f_w, N, Z)
\end{equation}

\hspace{4.7mm} 6. Update $w$ by using Adam to minimize loss.
}

\subsection{Notes on Optimal Transport}
There are multiple ways to calculate $\sigma$ from $N$ to $Z$. We employ the Python Optimal Transport library [17] which leverages the network simplex algorithm [18] in order to calculate the exact bijection. We also attempted using the Sinkhorn-Knopp algorithm to solve the entropic-regularized optimal transport problem, but the map $\sigma$ it provides is unhelpful due to its non-bijective nature. (In such a case, a neural network has a difficult time approximating it.)

\subsection{OTtrans vs OTgen}
We argue that OTgen is more like a latent space generator than a latent space transporter. In OTtrans, noise vectors are mapped to latent vectors nearby in order to minimize the total distance moved. Such a model has the advantage of only requiring to compute optimal transport once.

However, minimizing total distance moved is an artificial restriction and actually inhibits model performance. When generating an image with noise vector $n$, we don't necessarily want to generate the image whose latent space vector is closest to $n$, we simply want to generate a good image. OTgen is without this restriction and has more flexibility; the network $f_w$ is given the opportunity to transform any noise vectors before optimal transport is applied. Yet this requires computing optimal transport at every step.

\section{Experiments in Distribution Mapping}

We validate our algorithms with multiple experiments. First we show that OTgen and OTtrans both do well in generating points from lower dimensional distributions. We specifically use the two moons and concentric circles datasets as depicted below. Such experiments also give us a way to visualize how each algorithm functions. 

\begin{figure}[H]
    \centering
    \begin{minipage}[t]{0.45\textwidth}
        \centering
        \includegraphics[width=0.9\textwidth]{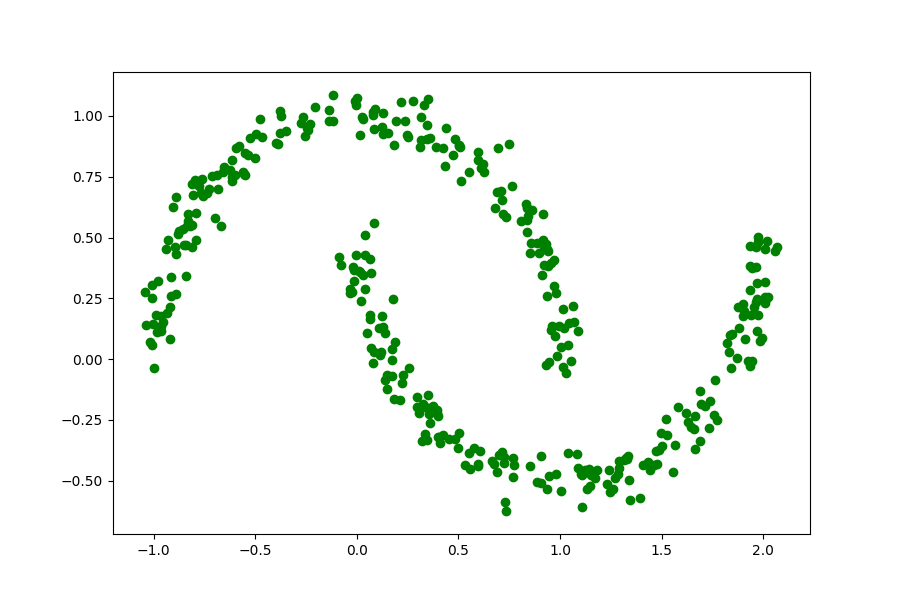}
    \end{minipage}
    \hspace{4mm}
    \begin{minipage}[t]{0.45\textwidth}
        \centering
        \includegraphics[width=0.9\textwidth]{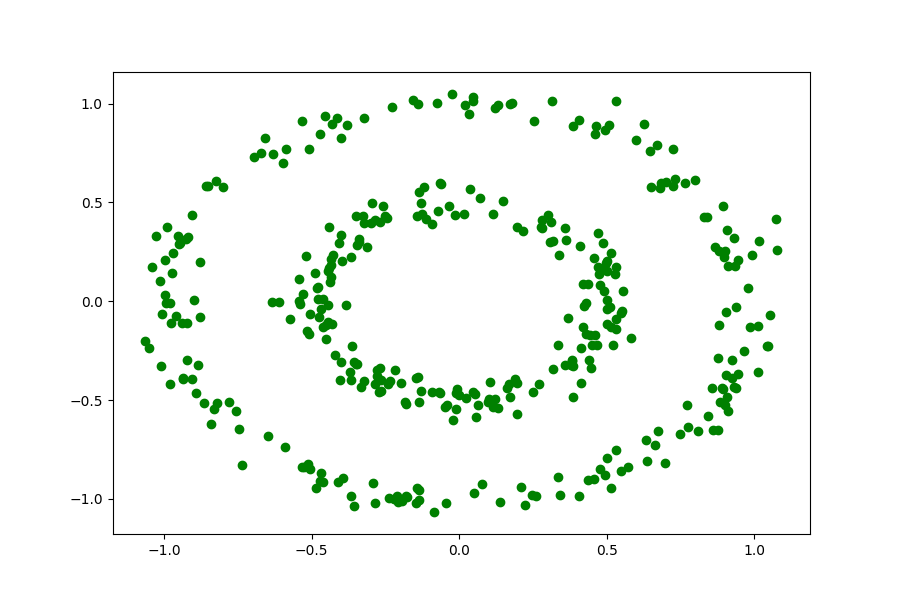}
    \end{minipage}
\end{figure}

In these experiments, the neural network architecture for OTgen and OTtrans is the same. It consists of 4 fully-connected layers of 512 neurons and a final layer of 2 neurons. Leaky ReLU [11] is used in between each layer. All networks are trained for a total of 10K steps, learning rate is set to 0.0003, and the prior noise distribution $\mathcal{N}$ is $\text{unif}(-1, 1)$. Batch size is set to 128. For OTgen, the diversity hyperparameter $\lambda$ is set to zero, as the dimension is small enough, we don't need to artificially increase diversity.

We calculate the divergence between the generated distribution and the real distribution by using optimal transport. This gives us a measure of the quality of the generated distribution and therefore the model; the lower the divergence, the closer the generated distribution is to the real distribution, and the better the model is. The divergence is simply the average distance when transporting optimally from the generated distribution to the real distribution. Again, we use the network simplex method to calculate the exact bijective mapping. The average distances are as follows:

\begin{center}
    \begin{tabular}{ |c|c|c|c| }  
        \hline
        Method & Moons & Circles \\
        \hline
        OTgen & 0.090 & 0.092 \\
        OTtrans & 0.086 & 0.075 \\
        Data & 0.070 & 0.071 \\
        \hline
    \end{tabular}
\end{center}

\subsection{Comparison to K-means}
For comparison, we also model the distributions with clusters. We apply K-means clustering to our data before approximating each cluster with a normal distribution. Intuitively, the more clusters there are, the more accurate the distribution will be approximated. Thus, if our models have high distribution modeling capabilities, they should be able to compare with an approximation with a high number of clusters.

\begin{center}
    \begin{tabular}{ |c|c|c|c| }  
        \hline
        Method & Moons & Circles \\
        \hline
        OTgen & 0.090 & 0.092 \\
        OTtrans & 0.086 & \textbf{0.075} \\
        Cluster (k=8) & 0.117 & 0.123 \\
        Cluster (k=16) & \textbf{0.084} & 0.090 \\
        Data & 0.070 & 0.071 \\
        \hline
    \end{tabular}
\end{center}

We find that both OTgen and OTtrans model each distribution better than the approximation with eight clusters and on-par with the approximation with sixteen clusters. Hence, it is shown that our models are very able in modeling lower dimensional data, often coming close to the optimal divergence. Examples of points generated by each model can be found in the appendix.

\subsection{OTgen's Training}

The two dimensional distributions also let us visualize OTgen's training and gain insight about its stochastic nature. OTgen's training process involves generating points and receiving "feedback" on the quality of each point. Yet this "feedback" is calculated based on the rest of the batch, which introduces some randomness. Consider the following two images:

\begin{figure}[H]
    \centering
    \begin{minipage}[t]{0.45\textwidth}
        \centering
        \includegraphics[width=0.9\textwidth]{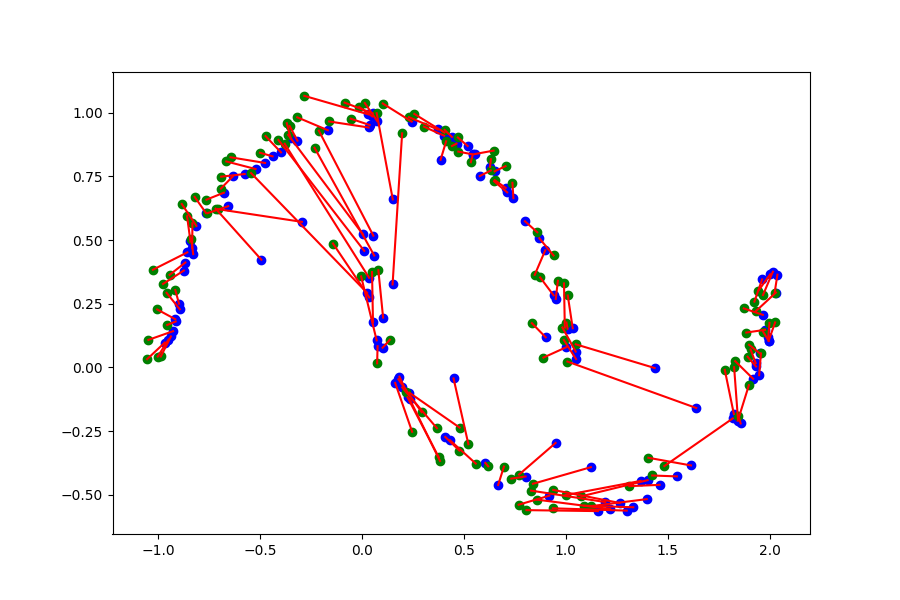}
    \end{minipage}
    \hspace{4mm}
    \begin{minipage}[t]{0.45\textwidth}
        \centering
        \includegraphics[width=0.9\textwidth]{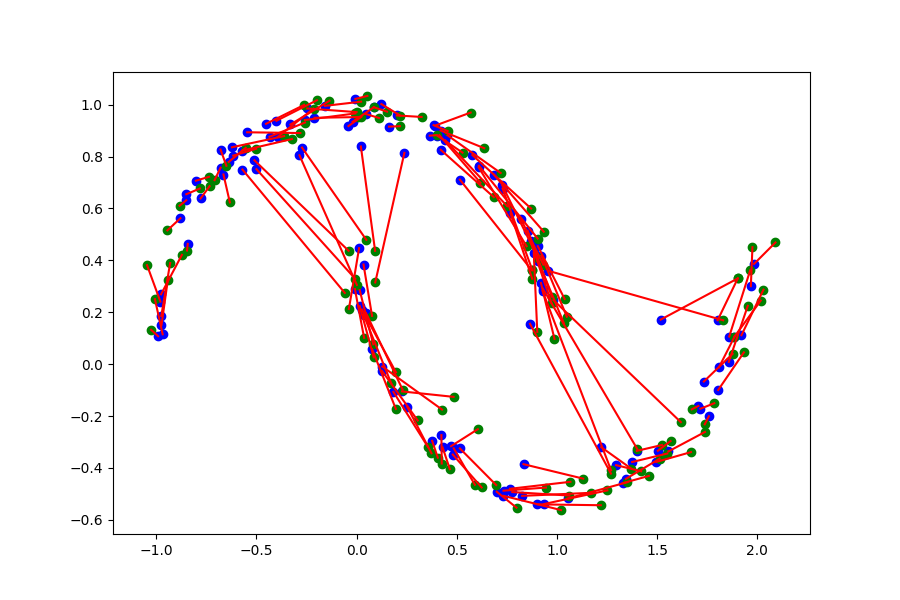}
    \end{minipage}
\end{figure}

These two images portray the feedback given on consecutive training steps. Blue dots are generated points, green dots are real data, and the red lines show a one to one mapping between the green and blue dots which minimizes overall distance travelled. Note that in this case, distance is L1 distance rather than traditional euclidean distance. The first image seems to indicate that the OTgen model is generating too many points in the bottom moon and that some should be mapped to the top moon. The second image tells the opposite story, that the model is generating too many points in the top moon and that some should be mapped to the bottom moon. The model isn't changing drastically between two consecutive training steps, so some of the feedback is wrong. However, on average the feedback provides useful information, so with enough training steps, the model converges.

\section{Experiments in Image Generation}
Next, we show that OTgen and OTtrans do well in generating points from a higher dimensional distribution, namely the latent space of an autoencoder trained either on the MNIST handwritten digits dataset [14], the Fashion MNIST clothing dataset [15], or the CelebA faces dataset [16]. The MNIST dataset of 28x28 black and white handwritten digits is widely regarded as the baseline dataset for many computer vision tasks, including image generation. The images are simple to generate and are without intricate patterns or gradients. Fashion MNIST is more difficult to generate than MNIST, as the clothing have different shades of gray and many difficult details, including shirt designs, stripes, frills, and gradients. Yet the Fashion MNIST dataset still is black and white and relatively small 28x28. The CelebA faces dataset is the most complicated dataset out of these three, with larger, colored images and faces showing different complex expressions. In our case we use the cropped CelebA images resized to 64x64x3. We compare against AE-OT, WAE-MMD, and VAE. 

\subsection{MNIST and Fashion MNIST}

For MNIST and Fashion MNIST, we did not use convolutional autoencoders but rather chose to only use fully connected layers. Each autoencoder's latent space dimension was set to eight. The OTgen mapping network consists of seven layers and its prior is set to $\text{unif}(-1,1)$. Both lambdas in WAE-MMD and AE-OT are set to 0.1, as is suggested in each respective paper. Batch size is set to 128. The diversity lamdba for OTgen was set to 0. Below the inception scores of the MNIST and Fashion MNIST images are shown (higher is better).

\begin{center}
    \begin{tabular}{ |c|c|c|c| }  
        \hline
        Method & MNIST & Fashion \\
        \hline
        True Images & 9.86 & 9.07 \\
        AE-OTgen & \textbf{9.52} & \textbf{7.90} \\
        AE-OTtrans & 9.19 & 7.45 \\
        AE-OT & 6.89 & 5.81 \\
        WAE-MMD & 7.46 & 5.97 \\ 
        VAE & 6.03 & 5.39 \\  
        GAN & 6.43 & 6.65 \\
        WGAN & 6.90 & 5.96 \\
        \hline
    \end{tabular}
\end{center}

In both MNIST and Fashion MNIST datasets, AE-OTgen comes the closest to the optimal inception score, with AE-OTtrans in second place. Both AE-OTgen and AE-OTtrans do substantially better than both the non-adversarial generators and adversarial generators such as GAN and WGAN. This demonstrates the capabilities of OTgen and OTtrans on natural latent distributions with low dimensions. In particular, our usage of optimal transport ensures diversity in that the model generates similar amounts of each class. Example images are shown in the appendix.

\subsection{CelebA}

For CelebA, we compare our models with the different non-adversarial models. The autoencoders in AE-OTgen, AE-OTtrans, AE-OT, VAE, and WAE all have the same architecture as in [2]. For AE-OTgen and AE-OTtrans, the mapping network is the same as MNIST. Finally, Batch size is set to 4096 to help increase the diversity of images. With a batch size of 4096, we ensure that our sampling from the latent distribution consistently matches the true latent distribution. AE-OT and WAE's lambdas are set to 0.1, AE-OT's diversity hyperparameter is set to 1. The Frechet Inception Distances on the CelebA dataset are shown below (lower is better).

\vspace{4mm}

\begin{center}
    \begin{tabular}{ |c|c|c|c| }  
        \hline
        Model & FID \\
        \hline
        AE-OTgen & \textbf{58.07} \\
        AE-OTtrans & 58.79 \\
        AE-OT & 106.96 \\
        WAE-MMD & 64.71 \\ 
        VAE & 59.85 \\  
        \hline
    \end{tabular}
\end{center}

From the FID scores, we see that AE-OTgen and AE-OTtrans again outperform WAE-MMD, VAE, and AE-OT. This shows its efficacy in modeling higher dimensional data and that it is state of the art in the field of non-adversarial generative modeling. As noted before, the images generated by VAE are very blurry. In contrast, though the images generated by WAE-MMD are very sharp, they often lack the facial structure. AE-OT is unable to handle the sparsity of the CelebA autoencoder's latent distribution and generates poor images. Example images pertaining to WAE-MMD, VAE, AE-OT, AE-OTtrans, and AE-OTgen are found in the appendix.

\subsection{AE-OTtrans vs AE-OTgen}

\begin{figure}[H]
    \centering
    \begin{minipage}{0.45\textwidth}
        \centering
        \includegraphics[width=0.7\textwidth]{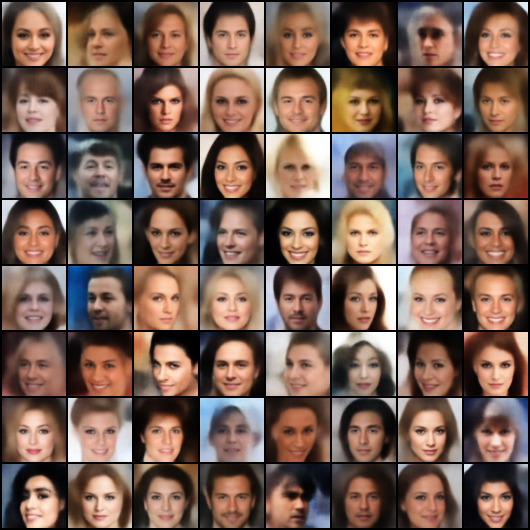} 
        \caption{AE-OTtrans}
    \end{minipage}
    \begin{minipage}{0.45\textwidth}
        \centering
        \includegraphics[width=0.7\textwidth]{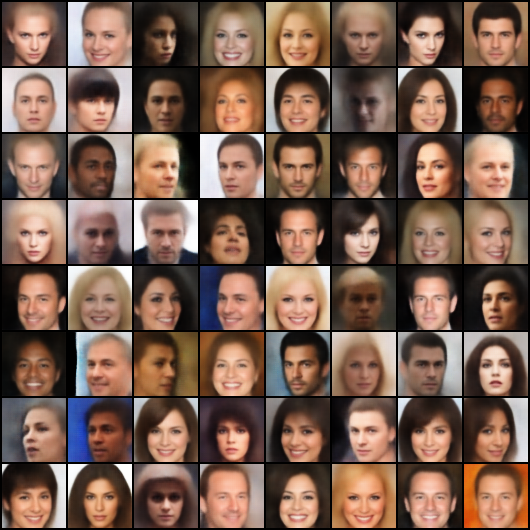}
        \caption{AE-OTgen}
    \end{minipage}
\end{figure}

Here, we compare the images generated by AE-OTtrans to the images generated by AE-OTgen. Both images are well structured, without major deformities. AE-OTtrans images are more diverse than AE-OTgen, with more varied face archetypes and backgrounds. Yet the faces generated by AE-OTgen are more sharp, albeit with less diversity. A similar pattern is seen when viewing the different model's interpolation.
Interpolation exists to ensure that each model doesn't simply memorize the different datapoints but instead can generate the whole distribution in a smooth fashion. The two model's interpolations are shown below:

\begin{figure}[H]
    \centering
    \begin{minipage}{0.45\textwidth}
        \centering
        \includegraphics[width=0.7\textwidth]{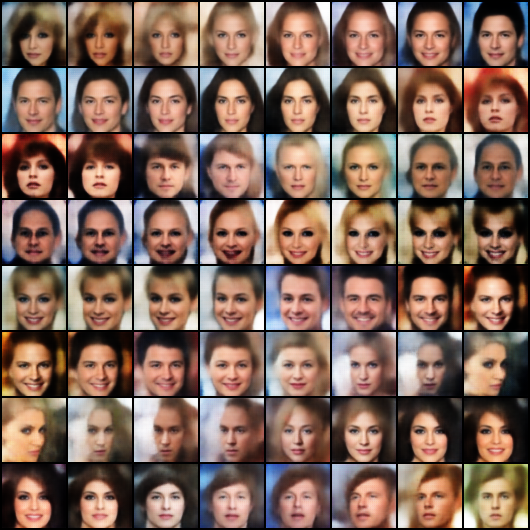} 
        \caption{Interpolation in AE-OTtrans}
    \end{minipage}
    \begin{minipage}{0.45\textwidth}
        \centering
        \includegraphics[width=0.7\textwidth]{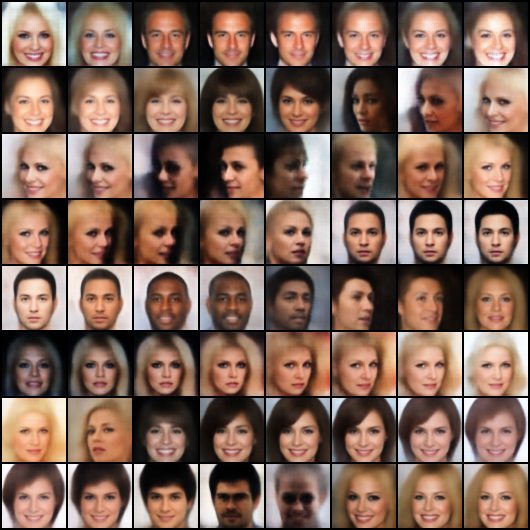}
        \caption{Interpolation in AE-OTgen}
    \end{minipage}
\end{figure}

\vspace{2mm}

AE-OTgen and AE-OTtrans have a significant difference with respect to interpolation. AE-OTtrans interpolation is more smooth, without many radical shifts in the image. Yet this comes at the cost of the image quality; the images are a bit more blurry and unrealistic. On the contrary, AE-OTgen interpolation is not as smooth, with more drastic shifts in the image, but most images are realistic and sharp. The suggests that AE-OTtrans interpolation is more natural and smooth, whereas in AE-OTgen, the interpolation is jumpy to ensure each transition image is realistic. 

%Overall we find that AE-OTtrans is a better generative model than AE-OTgen.

%\section{The Importance of Adversarial Training}
%By analyzing AE-OTgen's shortcomings, we can understand the effects of adversarial training. We argue that the limiting factor of AE-OTgen lies in its difficulty to generate diverse images, which stems from an inability of the L1Loss function to deal with uncertainty. From this perspective, the main value of adversarial training lies in its tendency to guess when uncertain rather than retreat to a default face.

%\subsection{AE-OTgen's lack of Diversity}

%AE-OTgen lacks diversity; though \_\%, \_\%, and \_\%, of the faces in the CelebA dataset have glasses, hats, and mustaches respectively, very few of these facial accessories are generated. Furthermore, if the diversity hyperparameter $\lambda$ is set to zero, the images generated seem to be the same face with only minor variations.

%[Image]

%Upon a first glance, it seems as if OTgen 

\section{Conclusion and Further Works}
In conclusion, we have proposed two models OTtrans and OTgen which do latent distribution mapping. These two models can be extended to AE-OTtrans and AE-OTgen which do high dimensional data generation without adversarial training. We've shown that OTtrans and OTgen are reasonable models when applied to two dimensional datasets, often outperforming the conventional modeling of distirbutions using clusters. Similarly, AE-OTtrans and AE-OTgen also do well, significantly outperforming VAE, WAE-MMD, and AE-OT on the MNIST and FashionMNIST and slightly outperforming the competing models on the CelebA dataset. Furthermore, AE-OTtrans and AE-OTgen outperform GANs on the MNIST and FashionMNIST dataset. Combined with the simpler training procedure of our new non-adversarial algorithm, this provides a compelling case to use AE-OTtrans or AE-OTgen for lower dimensional data generation over adversarial training. Future work will include an expanded theoretical analysis of AE-OTtrans and AE-OTgen as well as further improvements on the AE-OTtrans and AE-OTgen models.

\vspace{4mm}

\noindent\textbf{Acknowledgements}

\vspace{2mm}

\noindent The authors of this paper would like to acknowledge Zach Gaslowitz of Proof School for the many fruitful discussions along the way. We would also like to thank Dr. Mei Han from Ping An Technology for supporting and facilitating this project. Finally, we would like to thank all the students at Proof School who helped in proofreading the paper.

\vspace{6mm}

\noindent\large\textbf{References}

\small

\setlength\parindent{0pt}

\begin{enumerate}[label={[\arabic*]}]
  \item D. P. Kingma and M. Welling. Auto-encoding variational bayes. In \textit{ICLR}, 2014.
  \item I. Tolstikhin, O. Bousquet, S. Gelly, and B. Schoelkopf. Wasserstein auto-encoders. In \textit{ICLR}, 2018.
  \item H. Liu, Y. Guo, N. Lei, Z. Shu, S. T. Yau, D. Samaras, and X. Gu. Latent space optimal transport for generative models. \textit{arXiv preprint arXiv:1809.05964}, 2018
  \item N. Lei, K. Su, L. Cui, S.-T. Yau, and D. X. Gu. A geometric view of optimal transportation and generative model. \textit{arXiv preprint arXiv:1710.05488}, 2017.
  \item I. Goodfellow, J. Pouget-Abadie, M. Mirza, B. Xu, D. Warde-Farley, S. Ozair, A. Courville, and Y. Bengio. Generative adversarial nets. In \textit{Advances in Neural Information Processing Systems}, pages 2672-2680, 2014.
  \item M. Arjovsky, S. Chintala, and L. Bottou. Wasserstein GAN. \textit{arXiv preprint arXiv:1701.07875}, 2017.
  \item D. Berthelot, T. Schumm, and L. Metz. Began: boundary equilibrium generative adversarial networks. \textit{arXiv preprint arXiv:1703.10717}, 2017.
  \item N. Kodali, J. Abernethy, J. Hays, and Z. Kira. On convergence and stability of gans. \textit{arXiv preprint arXiv:1705.07215}, 2017.
  \item L. Mescheder, A. Geiger, and S. Nowozin. Which training methods for gans do actually converge? In \textit{International Conference on Machine Learning}, pages 3478-3487, 2018.
  \item A. Brock, J. Donahue, and K. Simonyan. Large scale GAN training for high fidelity natural image synthesis. In \textit{ICLR}, 2019.
  \item B. Xu, N. Wang, T. Chen, and M. Li. Empirical evaluation of rectified activations in convolutional network. In \textit{CoRR}, abs/1505.00853, 2015.
  \item S. Ioffe, C. Szegedy. Batch normalization: Accelerating deep network training by reducing internal covariate shift. In \textit{International Conference on Machine Learning}, 2015, pp. 448-456.
  \item M. Heusel, H. Ramsauer, T. Unterthiner, B. Nessler, and S. Hochreiter. Gans trained by a two time-scale update rule converge to a local nash equilibrium. In \textit{NIPS}, 6626-6637, 2017.
  \item Xiao, H., Rasul, K., and Vollgraf, R. Fashion-MNIST: A novel image dataset for benchmarking machine learning algorithms. In \textit{arXiv:1708.07747}, 2017
  \item Y. LeCun, L. Bottou, Y. Bengio, and P. Haffner. Gradient-based learning applied to document recognition. In \textit{Proceedings of the IEEE}, volume 86(11), pages 2278-2324, 1998.
  \item Z. Liu, P. Luo, X. Wang, and X. Tang. Deep learning face attributes in the wild. In \textit{ICCV}, 2015.
  \item R. Flamary, N. Courty. Python Optimal Transport. https://github.com/rflamary/POT, 2017.
  \item G. Peyre, M. Cuturi. Computational Optimal Transport. \textit{arXiv:1803.00567}, 2019.
\end{enumerate}

\newpage

\section{Appendix}

\subsection{Lower Dimensional Distributions}

\vspace{-7mm}

\begin{figure}[H]
    \centering
    \begin{minipage}{0.45\textwidth}
        \centering
        \includegraphics[width=0.9\textwidth]{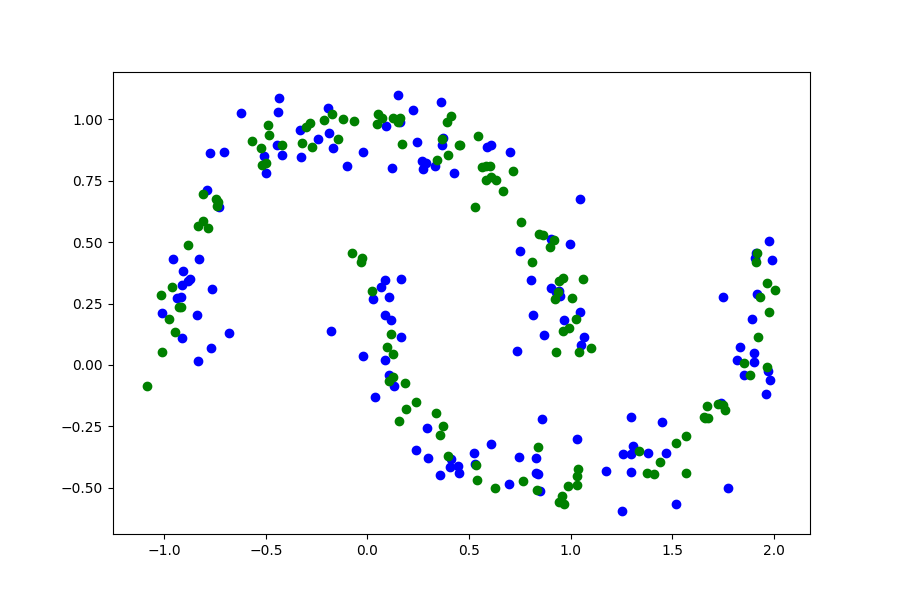} 
        \caption{Moons, Eight Cluster}
   \end{minipage}
    \begin{minipage}{0.45\textwidth}
        \centering
        \includegraphics[width=0.9\textwidth]{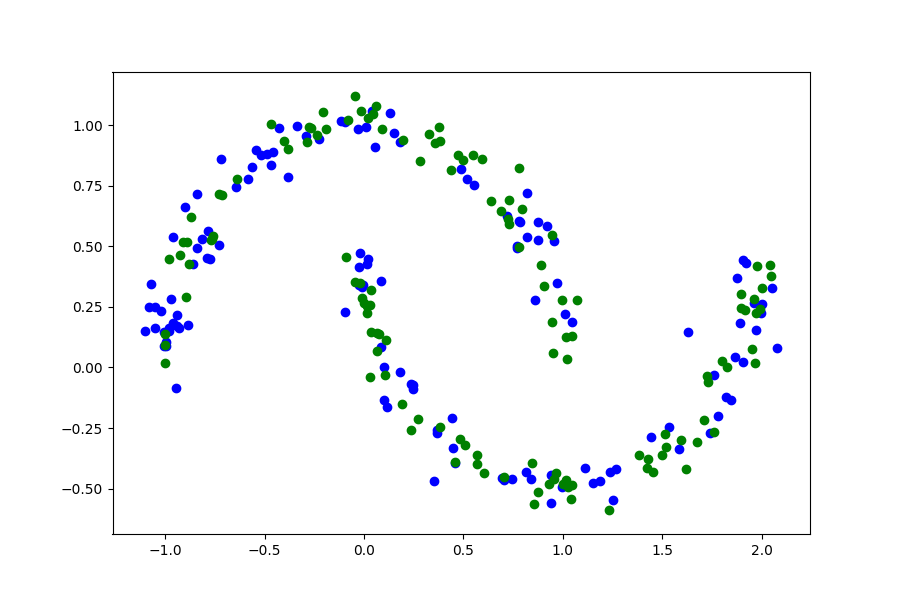}
        \caption{Moons, Sixteen Cluster}
    \end{minipage}
\end{figure}

\vspace{-8mm}

\begin{figure}[H]
    \centering
    \begin{minipage}{0.45\textwidth}
        \centering
        \includegraphics[width=0.9\textwidth]{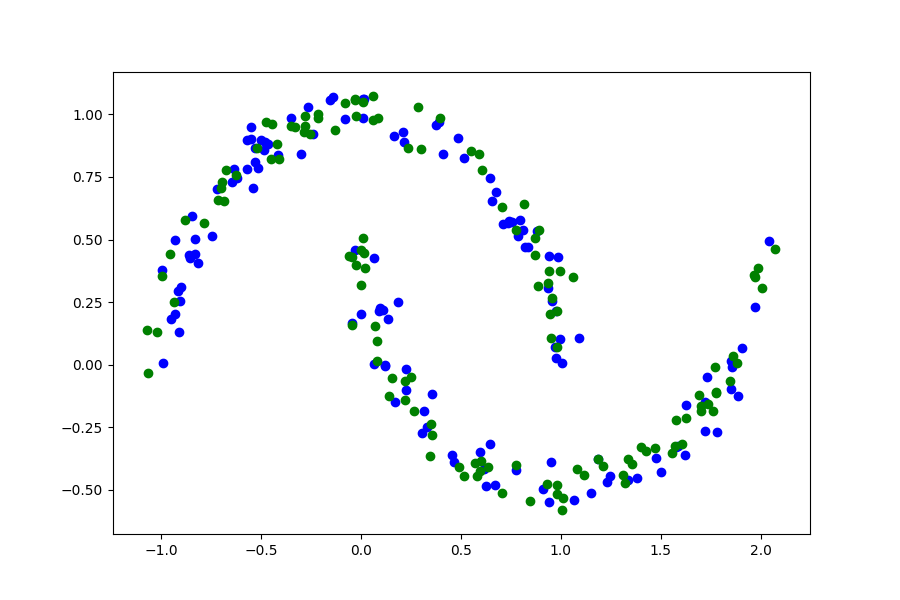}
        \caption{Moons, OTtrans}
    \end{minipage}
    \begin{minipage}{0.45\textwidth}
        \centering
        \includegraphics[width=0.9\textwidth]{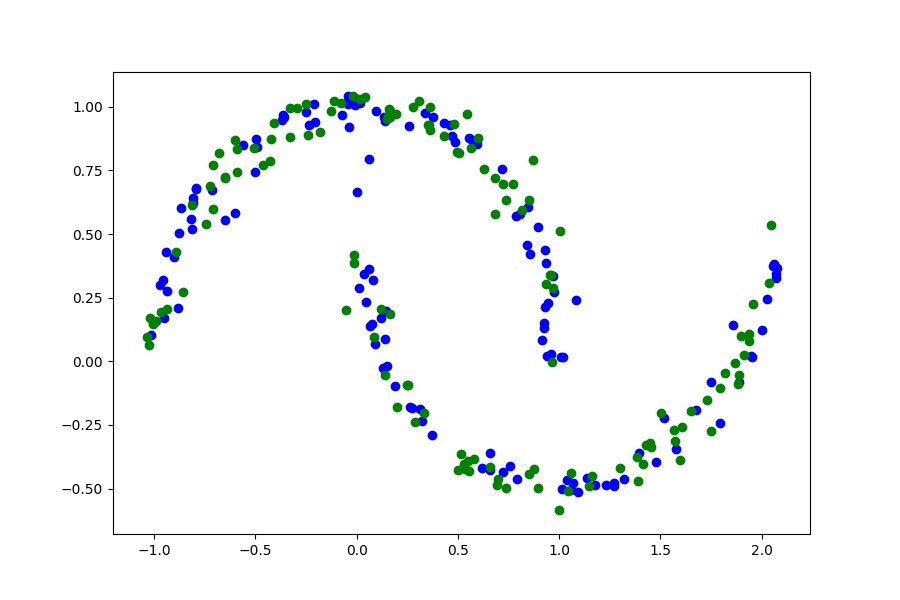}
        \caption{Moons, OTgen}
    \end{minipage}
\end{figure}

\begin{figure}[H]
    \centering
    \begin{minipage}{0.45\textwidth}
        \centering
        \includegraphics[width=0.9\textwidth]{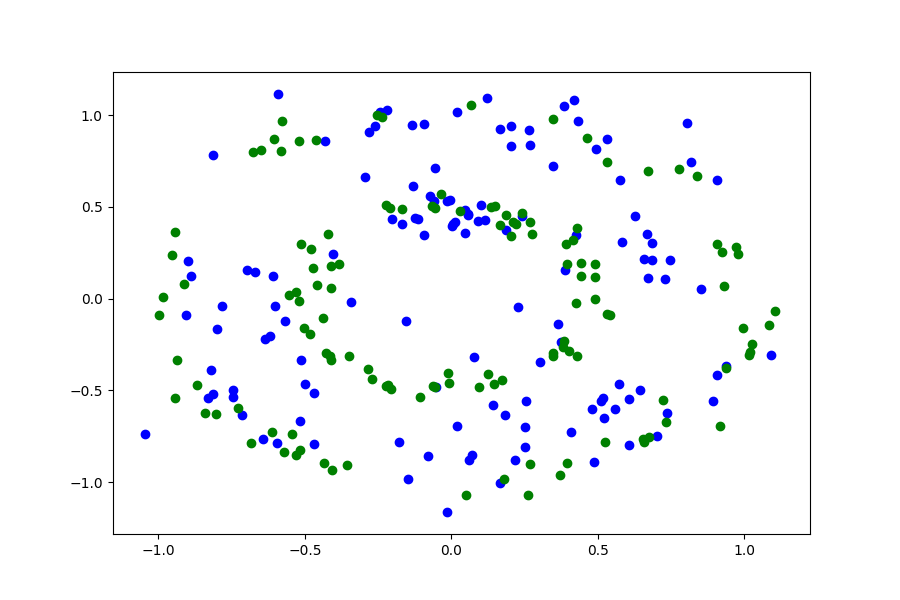}
        \caption{Circles, Eight Cluster}
   \end{minipage}
    \begin{minipage}{0.45\textwidth}
        \centering
        \includegraphics[width=0.9\textwidth]{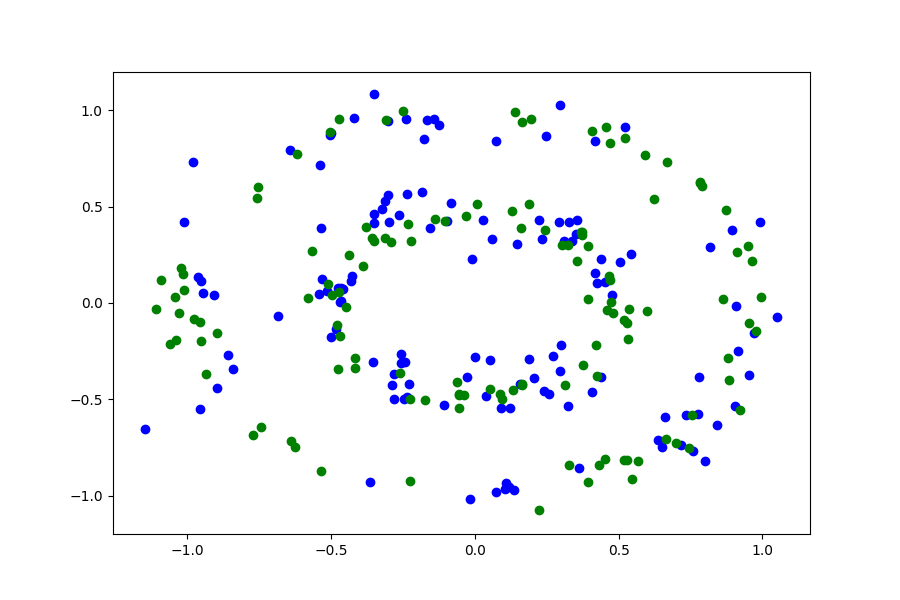}
        \caption{Circles, Sixteen Cluster}
    \end{minipage}
\end{figure}

\begin{figure}[H]
    \centering
    \begin{minipage}{0.45\textwidth}
        \centering
        \includegraphics[width=0.9\textwidth]{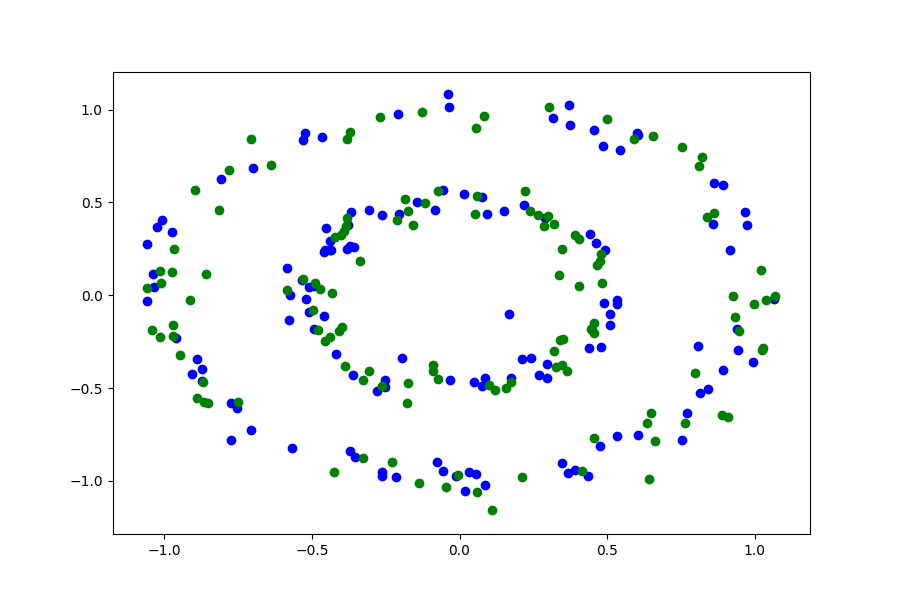}
        \caption{Circles, OTtrans}
    \end{minipage}
    \begin{minipage}{0.45\textwidth}
        \centering
        \includegraphics[width=0.9\textwidth]{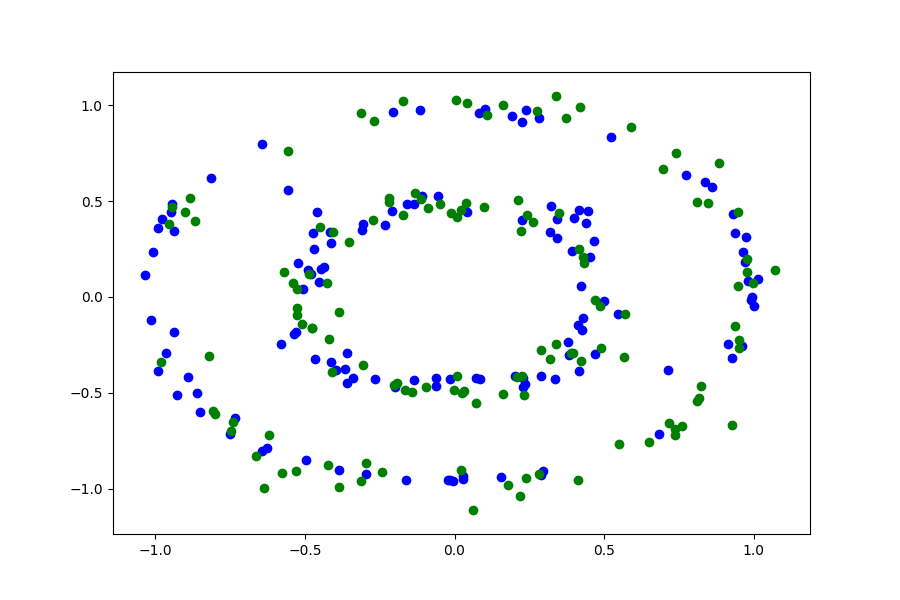}
        \caption{Circles, OTgen}
    \end{minipage}
\end{figure}

\subsection{MNIST Images}

\begin{figure}[H]
    \centering
    \begin{minipage}{0.3\textwidth}
        \centering
        \includegraphics[width=0.9\textwidth]{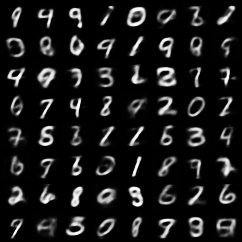} 
        \caption{VAE}
   \end{minipage}
    \begin{minipage}{0.3\textwidth}
        \centering
        \includegraphics[width=0.9\textwidth]{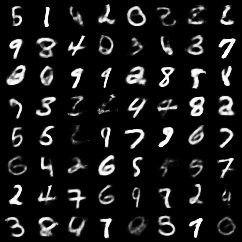} 
        \caption{WAE-MMD}
    \end{minipage}
    \begin{minipage}{0.3\textwidth}
        \centering
        \includegraphics[width=0.9\textwidth]{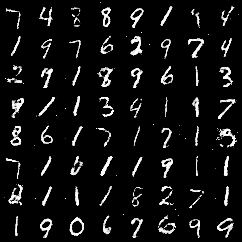}
        \caption{GAN}
    \end{minipage}
\end{figure}

\begin{figure}[H]
    \centering
    \begin{minipage}{0.3\textwidth}
        \centering
        \includegraphics[width=0.9\textwidth]{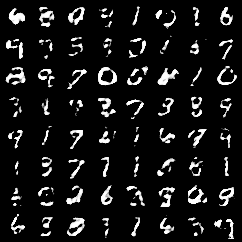} 
        \caption{WGAN}
   \end{minipage}
    \begin{minipage}{0.3\textwidth}
        \centering
        \includegraphics[width=0.9\textwidth]{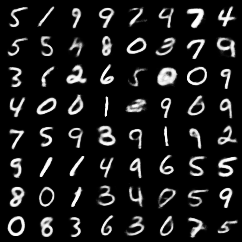} 
        \caption{AE-OTtrans}
    \end{minipage}
    \begin{minipage}{0.3\textwidth}
        \centering
        \includegraphics[width=0.9\textwidth]{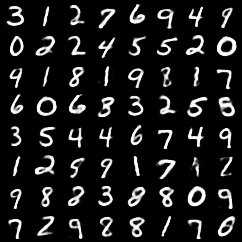}
        \caption{AE-OTgen}
    \end{minipage}
\end{figure}

\subsection{FashionMNIST Images}

\begin{figure}[H]
    \centering
    \begin{minipage}{0.3\textwidth}
        \centering
        \includegraphics[width=0.9\textwidth]{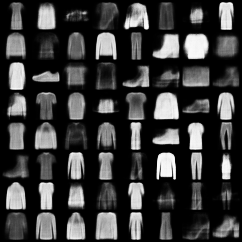} 
        \caption{VAE}
   \end{minipage}
    \begin{minipage}{0.3\textwidth}
        \centering
        \includegraphics[width=0.9\textwidth]{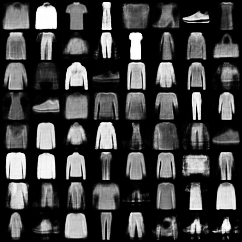} 
        \caption{WAE-MMD}
    \end{minipage}
    \begin{minipage}{0.3\textwidth}
        \centering
        \includegraphics[width=0.9\textwidth]{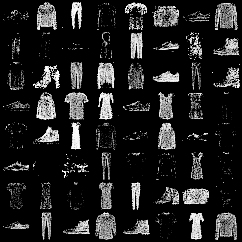}
        \caption{GAN}
    \end{minipage}
\end{figure}

\begin{figure}[H]
    \centering
    \begin{minipage}{0.3\textwidth}
        \centering
        \includegraphics[width=0.9\textwidth]{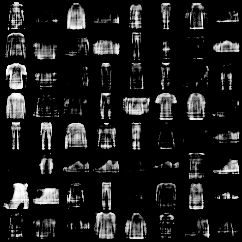} 
        \caption{WGAN}
   \end{minipage}
    \begin{minipage}{0.3\textwidth}
        \centering
        \includegraphics[width=0.9\textwidth]{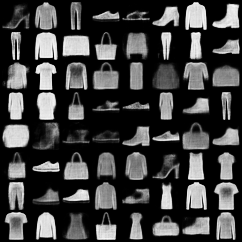} 
        \caption{AE-OTtrans}
    \end{minipage}
    \begin{minipage}{0.3\textwidth}
        \centering
        \includegraphics[width=0.9\textwidth]{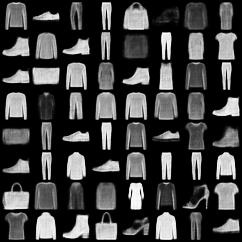}
        \caption{AE-OTgen}
    \end{minipage}
\end{figure}

\subsection{CelebA Images}

\begin{figure}[H]
    \centering
    \begin{minipage}{0.3\textwidth}
        \centering
        \includegraphics[width=0.9\textwidth]{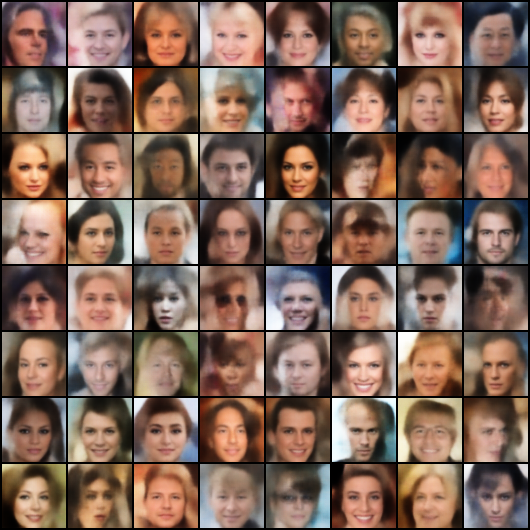} 
        \caption{VAE}
   \end{minipage}
    \begin{minipage}{0.3\textwidth}
        \centering
        \includegraphics[width=0.9\textwidth]{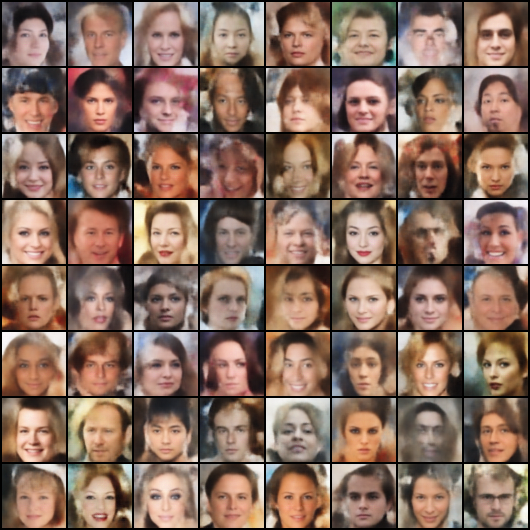} 
        \caption{WAE}
    \end{minipage}
    \begin{minipage}{0.3\textwidth}
        \centering
        \includegraphics[width=0.9\textwidth]{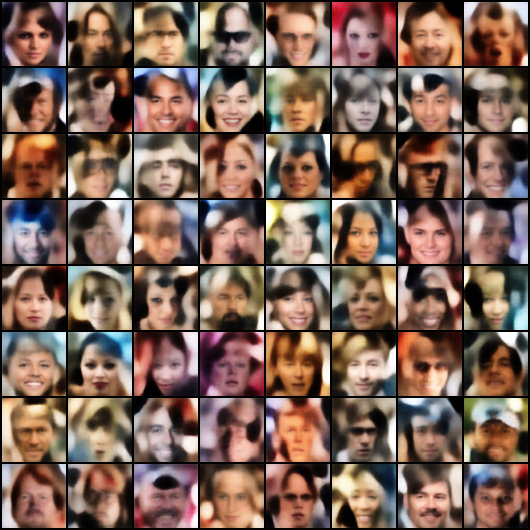}
        \caption{AE-OT}
    \end{minipage}
\end{figure}

\begin{figure}[H]
    \centering
    \begin{minipage}{0.45\textwidth}
        \centering
        \includegraphics[width=0.9\textwidth]{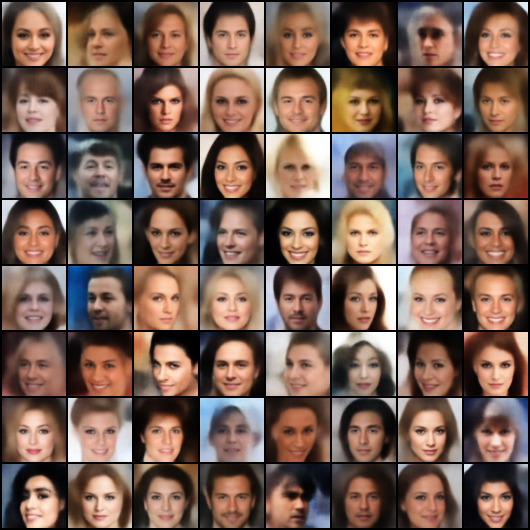} 
        \caption{AE-OTtrans}
    \end{minipage}
    \begin{minipage}{0.45\textwidth}
        \centering
        \includegraphics[width=0.9\textwidth]{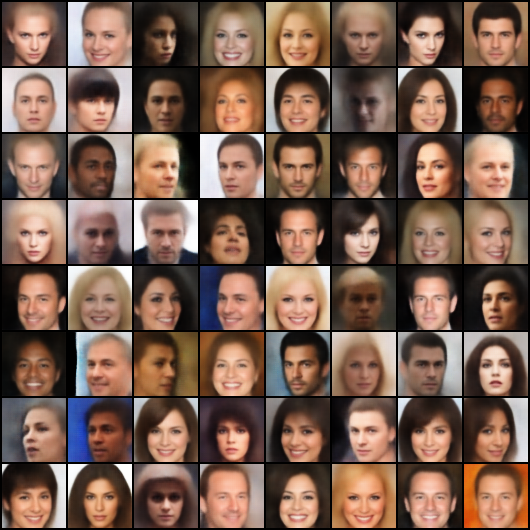}
        \caption{AE-OTgen}
    \end{minipage}
\end{figure}

\end{document}